\documentclass{article}
\usepackage{spconf,amsmath,graphicx,hyperref}
\usepackage{amssymb}
\usepackage{booktabs}
\usepackage{float}

\title{Improving Text-instance Alignment of Foreground Conditioned Out-painting via
Customized Concept Embedding}

\name{ Yihao Zhao$^{1,*}$, Xuan Han$^{1,*}$, Bin He$^{1,2}$, Mingyu You$^{1,2,\dagger}$ 
\thanks{$^{*}$Co-first authors. $^{\dagger}$Corresponding author: myyou@tongji.edu.cn} } 

\address{ $^{1}$College of Electronic and Information Engineering, Tongji University, Shanghai, China \\ $^{2}$State Key Laboratory of Autonomous Intelligent Unmanned Systems, \\ Frontiers Science Center for Intelligent Autonomous Systems, Ministry of Education, Shanghai, China }
         
\begin{document}
\ninept
\maketitle
\begin{abstract}
To showcase products, merchants often incur substantial costs creating high-quality display images. Foreground Conditioned Out-painting (FCO) meets this demand, allowing users to create desired backgrounds for foreground instances at a low cost by adjusting the text prompt. However, existing text-driven FCO methods exhibit critical flaws in their outputs, most notably the presence of artifacts, which refer to regions in the synthesized background that share the same semantics as the foreground instance. Such artifacts diminish the object's prominence and degrade image quality. We attribute the issue to the misalignment between the given instance and text-derived concept embeddings. To address this, we propose the \textbf{C}ustomized \textbf{C}oncept \textbf{E}mbedding Diffusion (CCE-Diffusion) framework. Its core is a CCE-Module to customize concept embeddings, bridging the gap between generic noun semantics (e.g., "car") and a specific visual instance (e.g., "a red sedan"). An Instance-Aware Loss guides the module's optimization, while a Semantic-Preserving Prompt Template prevents customized embeddings from distorting other words in the prompt. Both qualitative and quantitative evaluations demonstrate that CCE-Diffusion significantly reduces artifacts in the outputs. As a plug-and-play component, the CCE-Module can integrate with various FCO methods, enhancing their performance.  
\end{abstract}
\section{Introduction}
\label{sec:intro}
In fields like e-commerce, the high cost and slow pace of traditional workflows struggle to meet the demand for high-quality product imagery. To address this, text-driven Foreground Conditioned Out-painting (FCO) technology has emerged as a powerful, low-cost solution. Prominent FCO approaches~\cite{train_free_BLD, train_free_BD,T2I_SD, tuning_based_PowerPaint, Layerdiffuse, module_based_BrushNet, flux2024} address the task of synthesizing a coherent background from a foreground instance image and a text prompt, often by leveraging large-scale diffusion models~\cite{T2I_GLIDE, T2I_SD, T2I_DALLE, T2I_pixart}. These approaches follow two main strategies: training-free methods~\cite{train_free_BLD, train_free_BD, train_free_Null_text_inversion, train_free_inversion_1} that modify the standard denoising process of pre-trained models at inference time, and training-based methods~\cite{module_based_BrushNet, Layerdiffuse, module_based_Controlnet, tuning_based_1} that fine-tune the model to incorporate the instance image as an explicit condition.

\begin{figure}[t]
  \centering
  \includegraphics[width=\linewidth]{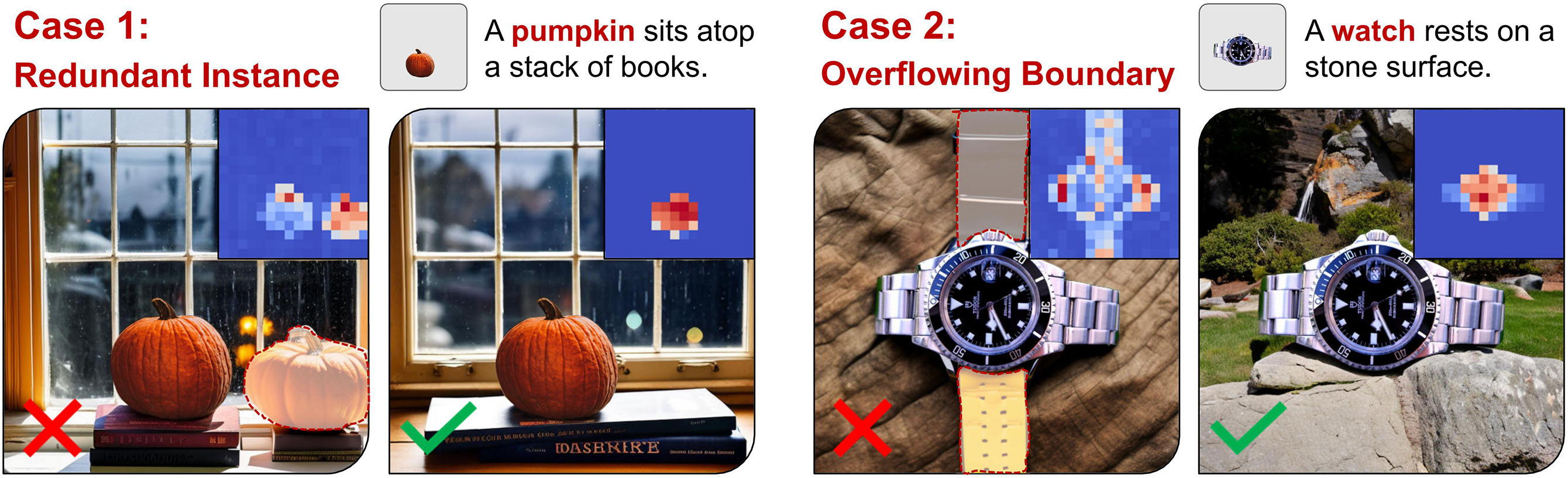}
  \vspace{-0.1cm}
  \caption{Two types of the artifacts phenomenon. Artifacts are indicated by red dashed lines.}
  \label{fig:introd}
\end{figure}

However, the existing text-driven FCO methods still exhibit critical flaws in their outputs, with artifacts~\cite{module_based_BrushNet, train_free_BLD} ranking among the most prominent issues. Artifacts refer to objects or regions in the synthesized background that have the same semantics as the foreground instance. Figure~\ref{fig:introd} illustrates two common artifact types: redundant Instance and overflowing boundary. The former will diminish the prominent presentation of the given object; in most commercial graphics, users do not want distracting elements appearing beside their products. The latter alter the appearance of the original products, making the generated image completely unusable.

To understand the formation of artifact, we need to revisit the principles of text-driven FCO methods. During the inference process, words in text prompts are first encoded as unique embeddings, which subsequently interact with image features in cross-attention layers~\cite{AttendAndExcite, e-semantic_diffusion_Prompt2Prompt}. Considering the text encoder~\cite{CLIP} is trained on large dataset, text embeddings tend to embed the generic concepts (e.g., "car"), whereas the given foreground image depicts an instance with a unique visual appearance (e.g., "a red sedan"). This discrepancy may cause misalignment between text embeddings and image features in the cross-attention layer. As shown in  Figure~\ref{fig:introd}, the concept embedding of the foreground object may be erroneously activated in the background regions, leading to artifacts in the final generated image. Therefore, suppressing artifacts in FCO requires enhancing the alignment between text embeddings and instance features.

In this paper, we propose a Customized Concept Embedding Diffusion (CCE-Diffusion) framework to address this problem. Its core component is the CCE-Module, which refines the generic concept embedding of an instance by conditioning it on the object's specific visual features. This process yields a customized concept embedding that is precisely aligned with the appearance of the given instance. The CCE-Module’s optimization is supervised by the Instance-Aware Loss. To prevent customized embeddings from distorting the semantics of other words in text prompts, we introduce a novel Semantic-Preserving Prompt Template and conduct a comprehensive analysis of its advantages over popular baseline templates. As a plug-and-play component, the CCE-Module can integrate with FCO methods. We validate its performance when combined with ControlNet~\cite{module_based_Controlnet}, BrushNet~\cite{module_based_BrushNet}, and BLD~\cite{train_free_BLD}, with significant artifact reduction in generated results.

The main contributions can be summarized as follow:

\begin{itemize}
\item We provide a theoretical analysis of the root causes of artifacts in FCO, thereby establishing a clear link between the misalignment of text prompt with instance visual features and the resulting artifact formation.
\item We propose a plug-and-play CCE-Module that customizes concept embeddings to improve text-instance alignment and suppress artifacts. Its optimization is guided by a new proposed Instance-Aware Loss.
\item We design a Semantic-Preserving Prompt Template to prevent customized embeddings from distorting the semantics of other words in prompt, thereby maintaining the overall text-image semantic consistency.
\end{itemize}

\section{Method}
\label{sec:Method}
Given an instance image $c \in \mathbb{R}^{h \times w \times ch}$, the FCO model generates an image $x$ guided by a text prompt $y$. $x$ should be artifact-free and conform to $y$. We adopt Stable Diffusion~\cite{T2I_SD} as our base model which is trained with the $\epsilon-prediction$ objective, formulated as:

\begin{equation}
    \mathcal{L}_{\epsilon-pred} = \mathbb{E}_{(z_0, y), t} \left[ \| \epsilon_t - \Phi(z_t, y, t) \|_2^2 \right]
    \label{eq:prediction}
\end{equation}

\begin{figure}[t]
  \centering
  \includegraphics[width=\linewidth]{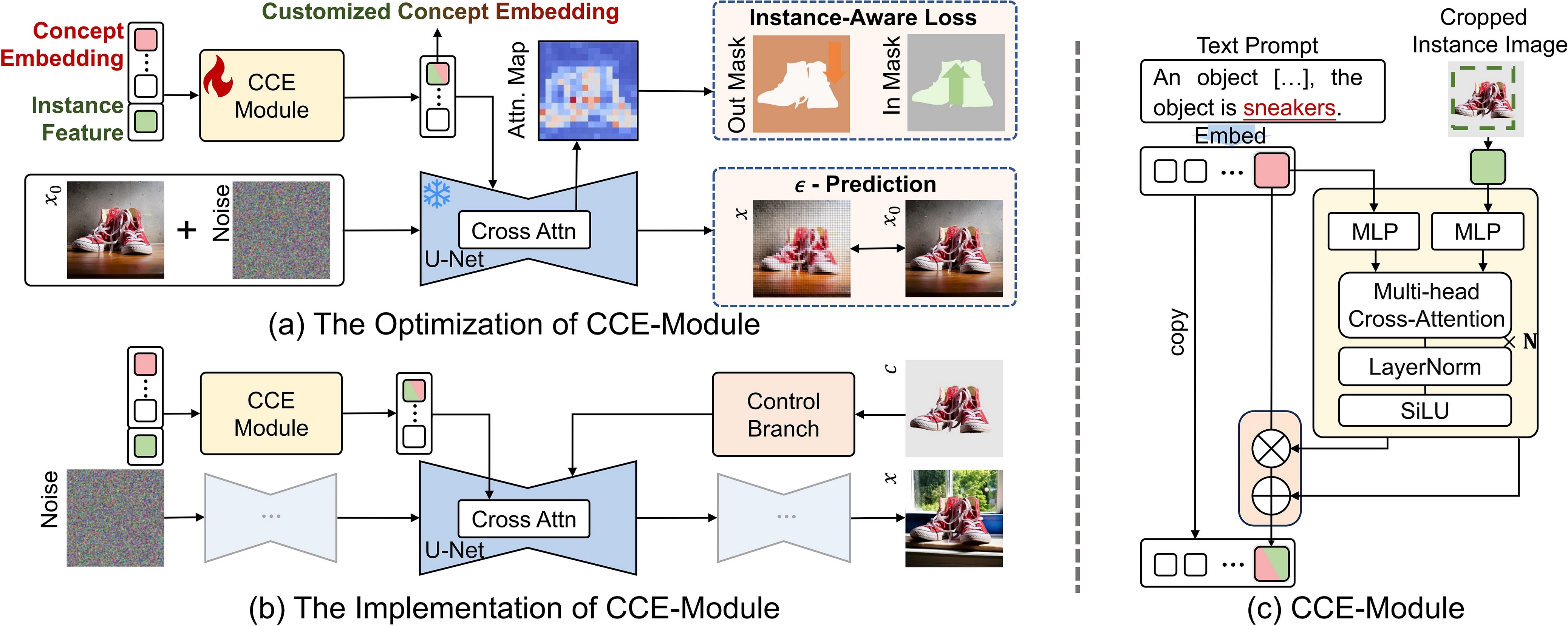}
  \caption{An overview of our proposed CCE-Module. (a) The optimization process with the Instance-Aware Loss. (b) The implementation details where the module acts as a plug-and-play component. (c) The detailed CCE-Module architecture.}
  \label{fig:method_1}
\end{figure}

Here, $z_t$ is a noisy latent created by adding Gaussian noise $\epsilon_t$ to the clean data latent $z_0$ at timestep $t$. The network $\Phi$ is trained to accurately predict the original added noise $\epsilon_t$ from $z_t$, conditioned on both the timestep $t$ and the text prompt $y$.

To eliminate artifacts, we propose the Customized Concept Embedding Diffusion (CCE-Diffusion) framework, as shown in Figure~\ref{fig:method_1}. It introduces a CCE-Module, trained with an Instance-Aware Loss from cross-attention maps to generate customized concept embeddings, and a novel Semantic-Preserving Prompt Template to ensure alignment with the input prompt $y$. The trained CCE-Module is designed as a plug-and-play component, allowing for easy integration with various FCO methods by inserting it before the text encoder~\cite{CLIP}, while keeping the base model frozen. Figure~\ref{fig:method_1}~(b) illustrates our best-performing setup, where the CCE-Module is integrated with a pre-trained ControlNet-based~\cite{module_based_Controlnet} branch, which remains independent of the optimization process shown in Figure~\ref{fig:method_1}~(a). 

\subsection{Customized Concept Embedding}
\label{subsec:CCE_Module}

In the FCO task, the misalignment between the prompt $y$ and foreground $c$ often leads to artifacts by causing high activations in background regions where foreground objects erroneously appear. To resolve this, we introduce the CCE-Module (Figure~\ref{fig:method_1}~(c)) to correct this misalignment at its root.

The process begins with two inputs: the text prompt $y$ and the cropped instance image $c$. First, the prompt is processed by a tokenizer and an embedding layer to produce word embeddings, from which we extract the instance embedding $e_{ins}$ (e.g., for "sneakers"). Concurrently, a CLIP image encoder extracts a visual feature vector from the instance image $c$, which is tightly cropped to the object's bounding box to remove the uniform background. As shown in Figure~\ref{fig:method_1}~(c), both $e_{ins}$ and the visual features are fed into our CCE-Module. The module, which consists of stacked layers including MLPs and multi-head cross-attention, processes these inputs to predict a weight vector $w$ and a bias vector $b$. These are used to perform an affine transformation on the original instance embedding, yielding the customized concept embedding via $e_{cc} = w \odot e_{ins} + b$. Finally, $e_{cc}$ replaces $e_{ins}$ in the token sequence, which is then passed to the text transformer~\cite{CLIP} to obtain the final text conditioning. 

\subsection{Instance-Aware Loss}
\label{subsec:IAL}
Although the CCE-Module refines the instance embedding, inaccuracies can still cause it to resemble background semantics. This leads to high activations in the background regions of cross-attention maps, resulting in artifacts or a degradation in overall image quality. To mitigate this, we propose an Instance-Aware Loss to further supervise the training of $e_{cc}$ for higher accuracy.

\begin{figure}[t]
  \centering
   \includegraphics[width=\linewidth]{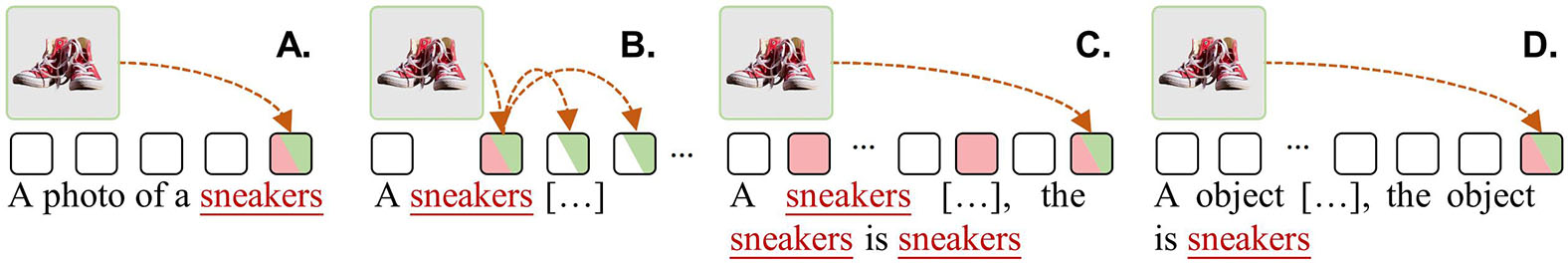}
   \caption{The process by which four different templates influence the semantics of text prompts. Red denotes generic concept embeddings, while green represents the visual features of an instance.}
   \label{fig:method_2}
\end{figure}

Building on prior work~\cite{AttendAndExcite, e-semantic_diffusion_Prompt2Prompt, e-stylistic_diffusion_4} that uses cross-attention maps for spatial guidance, we extract the maps $A_{cc}$ for our embedding $e_{cc}$ from the U-Net's cross-attention layers. We find that using maps with a resolution of $16\times 16$ yields satisfactory results. These maps are then averaged across all layers to produce a single attention map $\bar{A}_{cc}$. Given a foreground mask $m$ derived from the instance image, the loss is formulated as:

\begin{equation}
  \mathcal{L}_{ia} = \frac{1}{N} \sum \left( \bar{A}_{cc} \odot (1 - m) \right) - \frac{1}{N} \sum \left( \bar{A}_{cc} \odot m \right)
\end{equation}

where $N$ is the number of elements in the attention map $\bar{A}_{cc}$. This objective explicitly penalizes the activation of $e_{cc}$ in the background region (where mask $m$ is 0) while rewarding its activation in the foreground. This process enhances the model's spatial awareness, ensuring $e_{cc}$ is more aligned with the instance. The final training objective is a weighted sum:

\begin{equation}
  \mathcal{L} = \mathcal{L}_{\epsilon-pred} + \lambda _{ia} \mathcal{L}_{ia}
\end{equation}

where $\lambda_{ia}$ is the weight to balance the two loss terms, which is set to 0.01 based on our experiments.

\begin{figure}[t]
  \centering
  \includegraphics[width=\linewidth]{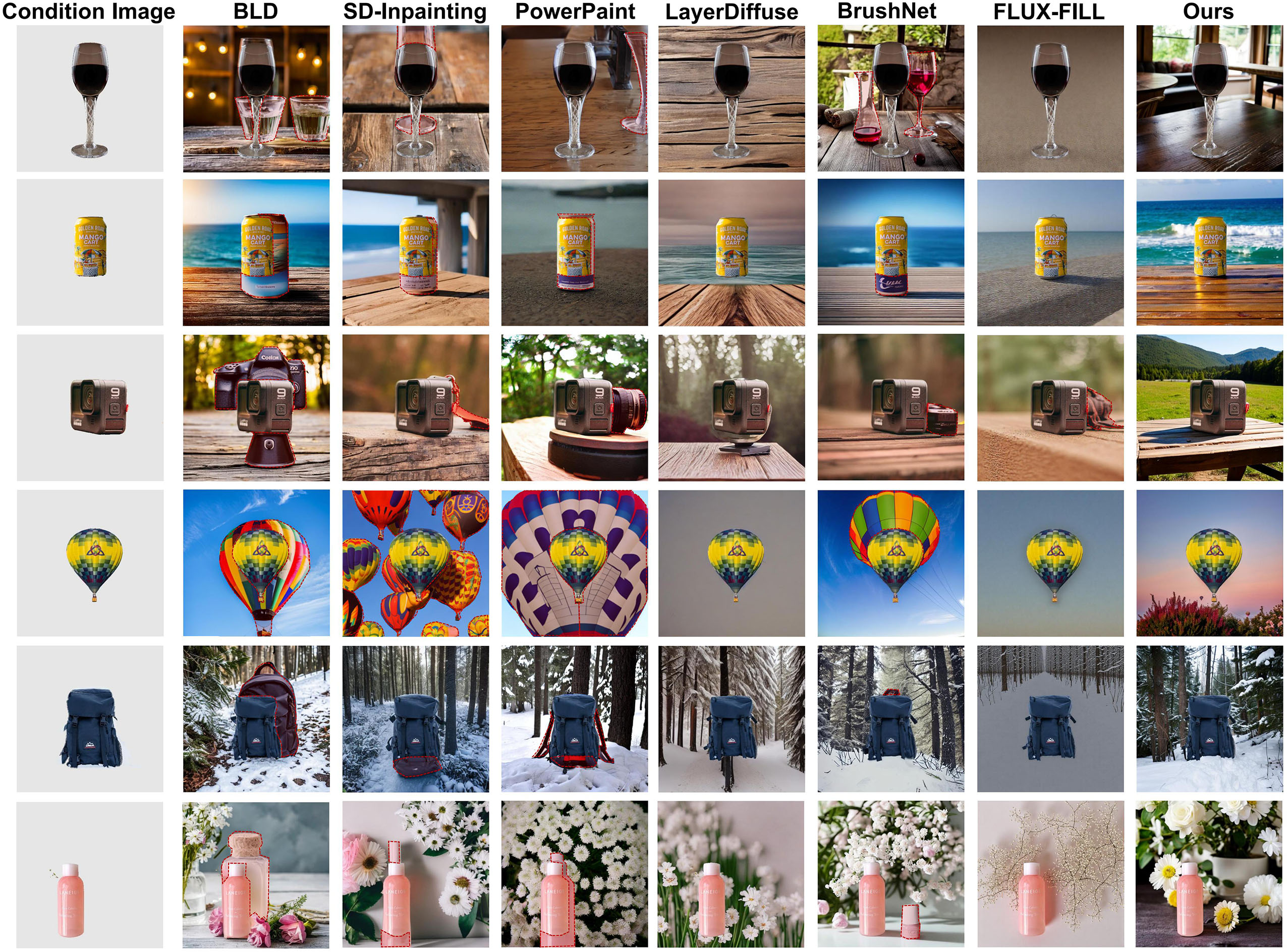}
  \caption{Qualitative comparison demonstrating that our method generates fewer artifacts (highlighted by red dashed lines) and achieves superior visual quality. For a fair comparison, the ground truth foreground is pasted into all results.}
  \label{fig:comparison}
\end{figure}

\begin{figure}[t]
  \centering
   \includegraphics[width=\linewidth]{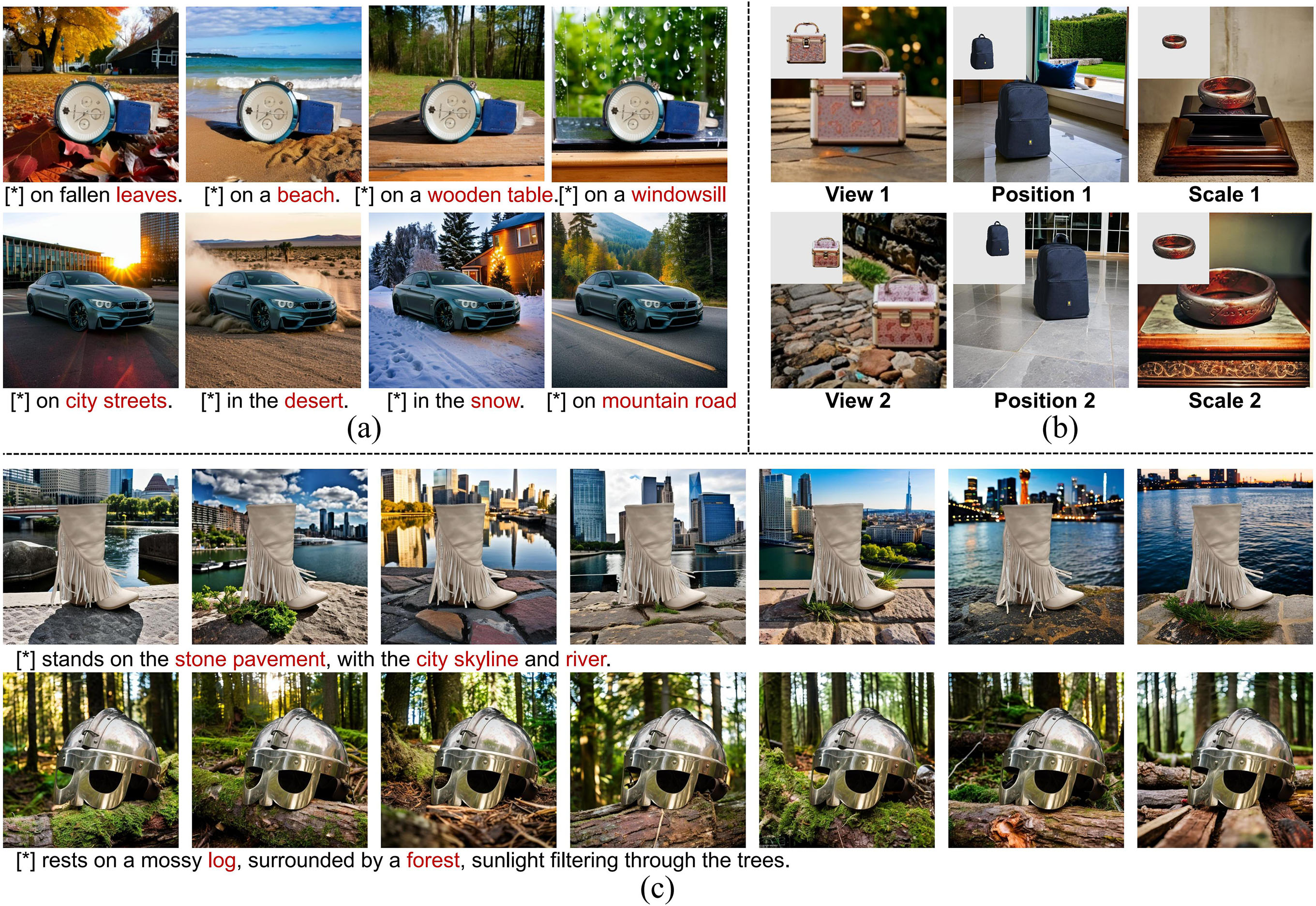}

  \caption{Qualitative results of our method demonstrating: (a) controllability over different backgrounds, (b) robustness to object variations, and (c) diversity in generated results.}
   \label{fig:multicase}
\end{figure}

\subsection{Semantic-Preserving Prompt Template}
\label{subsec:templates}

To effectively integrate our customized concept embedding ($e_{cc}$), we analyze various prompt structures, as illustrated in Figure~\ref{fig:method_2}. Using "sneakers" as an example, the figure shows how the $e_{cc}$ influences other words in the prompt. This behavior is governed by the CLIP~\cite{CLIP} text encoder's causal attention mechanism, expressed as $A = \text{softmax}\left(\frac{QK^T + L}{\sqrt{d}}\right)$, where the lower-triangular matrix $L$ ensures a token is only influenced by preceding tokens. Consequently, the position of $e_{cc}$ within the prompt sequence becomes critical. For instance, placing $e_{cc}$ at the beginning of the prompt would cause it to unduly influence the semantics of all subsequent tokens, leading to a degradation in the quality of the generated background.

Our analysis reveals issues with existing templates. \textbf{Template A}~\cite{Textualinversion} ("A photo of a $<$concept$>$") results in uncontrollable backgrounds, conflicting with FCO tasks. For a fair experimental comparison, we test this approach using \textbf{Template B}~\cite{T2I_SD} at inference. However, Template B, which represents common human-readable text prompts, risks semantic distortion if $e_{cc}$ is front-positioned. \textbf{Template C}~\cite{Subject-Diffusion, BLIP-Diffusion} ("..., the $<$concept$>$ is $<$$e_{cc}$$>$") places $e_{cc}$ safely at the end, but the separate concept word $<$concept$>$ creates a semantic conflict that can induce artifacts.

Based on the above analysis, we propose the \textbf{Semantic Preserving Prompt Template (Template D)}. We modify Template C by replacing the specific $<$concept$>$ with a generic placeholder  "object", which resolves the semantic conflict while retaining the safe end-positioning for $e_{cc}$. Experiments demonstrate the proposed template’s effectiveness for the final result.

\begin{table}[t]
  \centering
  \begin{tabular}{@{}lcccccc@{}}
    \toprule
    Method & \multicolumn{1}{c}{Artifacts} & \multicolumn{4}{c}{Image Quality} \\
    \cmidrule(lr){2-2} \cmidrule(lr){3-6}
    & SAM $\downarrow$  & TI $\uparrow$ & IR $\uparrow$ & LA $\uparrow$ & US $\uparrow$ \\
    \midrule
    BLD~\cite{train_free_BLD} & 0.986  & 26.06 & 0.18 & 5.04 & 1.35 \\
    SD-Inpaint.~\cite{T2I_SD} & 0.317  & 26.26 & 0.35 & 5.19 & 3.41 \\
    PowerPaint~\cite{tuning_based_PowerPaint} & 0.820 & 26.13 & 0.34 & 5.05 & 2.66 \\
    LayerDiff.~\cite{Layerdiffuse} & \textbf{0.046} & 26.15 & 0.22 & 5.27 & 3.78 \\
    BrushNet~\cite{module_based_BrushNet} & 0.190  & \textbf{26.62} & 0.60 & 5.31 & 4.27 \\
    FLUX-FILL$^\dagger$~\cite{flux2024}& 0.095 & 26.00 &0.52 & 5.03 & 4.18 \\
    \midrule
    \textbf{Ours (BLD)} & 0.543  & 26.18 & 0.37 & 5.17 & 1.85 \\
    \textbf{Ours (B-Net)} & 0.109  & \underline{26.44} & \underline{0.62} & \textbf{5.35} & \underline{4.31} \\
    \textbf{Ours} & \underline{0.086}  & 26.37 & \textbf{0.63} & \underline{5.33} & \textbf{4.45} \\
    \bottomrule
  \end{tabular}
  \caption{Comparison of different methods across various metrics. The items in bold are ranked first, and those with underlines are ranked second. $^\dagger$Indicates a model with an order of magnitude more parameters.}
  \label{tab:comparison}
\end{table}

\section{Experiments}
\label{sec:Experiments}

\subsection{Experimental Settings}
\noindent
\textbf{Dataset} Our training dataset, derived from OpenImage v7~\cite{OpenImage}, contains 580K images with 800K high-quality instance masks, each paired with a corresponding instance-related word. For evaluation, we curated a test set of 250 cases (100 from OpenImage validation, 150 web-sourced), each with high-quality manual annotations for masks, instance words, and captions, accompanied by 4 image samples per case to facilitate comprehensive model evaluation.

\noindent
\textbf{Implementation Details} Our work builds upon Stable Diffusion v2.1, with an additional version trained on Stable Diffusion v1.5 for integration with BrushNet~\cite{module_based_BrushNet}. We train all models at a $512\times512$ resolution for 340k steps using the AdamW~\cite{AdamW} optimizer, with a learning rate of $1\times 10^{-4}$ and a global batch size of 8. For inference, all images are generated using the DDIM sampler~\cite{DDIM} with 100 steps and a guidance scale of 7.0 for classifier-free guidance.

\noindent
\textbf{Evaluation Metrics} We evaluate our method using two categories of metrics. To evaluate the severity of artifacts, we use the \textbf{SAM} score ($\downarrow$)~\cite{SAM, GDINO}, which measures the discrepancy between generated instance masks and the ground truth. For overall image quality, we employ three automated metrics: text-image semantic similarity (\textbf{TI}, $\uparrow$)~\cite{CLIP}, a human preference score (\textbf{IR}, $\uparrow$)~\cite{ImageReward}, and an aesthetic score (\textbf{LA}, $\uparrow$)~\cite{Laion}. Furthermore, we conduct a User Study \textbf{(US), $\uparrow$}, in which 35 participants rated 30 cases on a 1-to-5 scale, with each case providing four generated images for evaluation. 

\begin{table}[t]
  \centering
  \begin{tabular}{@{}lccccc@{}} 
    \toprule
    Method & \multicolumn{1}{c}{Artifacts} & \multicolumn{4}{c}{Image Quality} \\ 
    \cmidrule(lr){2-2} \cmidrule(lr){3-6} 
    & SAM $\downarrow$ & TI $\uparrow$ & IR $\uparrow$ & LA $\uparrow$ & US $\uparrow$ \\ 
    \midrule
    Baseline~\cite{module_based_Controlnet} & 0.224 & 26.24 & 0.54 & 5.29 & 4.33 \\
    + CCE-Module & 0.115 & 25.41 & 0.45 & 5.14 & 4.20 \\
    \textbf{Ours} & 0.086 & \textbf{26.37} & \textbf{0.63} & \textbf{5.33} & \textbf{4.45} \\
    \bottomrule
  \end{tabular}
  \caption{Quantitative results for the ablation study of our proposed components (CCE-Module and Instance-Aware Loss).}
  \label{tab:ablation_study}
\end{table}

\begin{table}[t]
  \centering
  \begin{tabular}{@{}lccccc@{}} 
    \toprule
    Method & \multicolumn{1}{c}{Artifacts} & \multicolumn{4}{c}{Image Quality} \\ 
    \cmidrule(lr){2-2} \cmidrule(lr){3-6} 
    & SAM $\downarrow$ & TI $\uparrow$ & IR $\uparrow$ & LA $\uparrow$ & US $\uparrow$ \\ 
    \midrule
    Template A & 0.077 & 25.97 & 0.41 & 5.22 & 4.05 \\
    Template B & \textbf{0.069} & 25.83 & 0.42 & 4.97 & 3.98 \\
    Template C & 0.168 & 26.17 & 0.62 & 5.26 & 4.13 \\
    \textbf{Ours} & 0.086 & \textbf{26.37} & \textbf{0.63} & \textbf{5.33} & \textbf{4.45} \\
    \bottomrule
  \end{tabular}
  \caption{Quantitative comparison of different prompt templates.}
  \label{tab:template_comparison}
\end{table}

\subsection{Comparison with Existing Methods}

For evaluation, we compare our method against several state-of-the-art FCO approaches, including training-free~\cite{train_free_BLD} and training-based~\cite{T2I_SD, tuning_based_PowerPaint, Layerdiffuse, module_based_BrushNet, flux2024} methods. The quantitative and qualitative results are presented in Table~\ref{tab:comparison} and Figure~\ref{fig:comparison}. Our model is presented in three variants by combining it with BLD~\cite{train_free_BLD}, BrushNet~\cite{module_based_BrushNet}, and ControlNet~\cite{ module_based_Controlnet}, denoted as \textbf{Ours (BLD)}, \textbf{Ours (B-Net)}, and \textbf{Ours}.

As shown in Table~\ref{tab:comparison}, our method (\textbf{Ours}) achieves a superior balance between artifact mitigation and overall image quality. The effectiveness of our framework is further demonstrated as \textbf{Ours (BLD)} significantly enhances BLD across all metrics, and \textbf{Ours (B-Net)} reduces BrushNet's artifacts while maintaining comparable quality. 

\noindent
\textbf{Comparison with Training-free Methods:} As shown in Table~\ref{tab:comparison}, the method BLD~\cite{ train_free_BLD} performs poorly, with severe artifacts leading to a high SAM score of 0.986. Its simple "paste" mechanism in latent space lacks foreground-background interaction, restricting its use to minor inpainting tasks like filling small holes.

\noindent \textbf{Comparison with Training-based Methods:} Training-based methods generally underperform our approach as they largely overlook the critical misalignment between concept embeddings and instance features. Even the leading methods exhibit this trade-off. While LayerDiffuse~\cite{Layerdiffuse} suppresses artifacts, it does so at the cost of realism (e.g., the floating "can" in Figure~\ref{fig:comparison}). Conversely, while BrushNet~\cite{module_based_BrushNet} achieves high text similarity with high CLIP score, its generated images remain unsuitable for the FCO task due to noticeable artifacts.

\noindent
\textbf{Comparison with FLUX-FILL:} Notably, FLUX-FILL~\cite{flux2024}, a state-of-the-art model with 12B parameters (far exceeding our 0.8B Stable Diffusion v2.1), fails to surpass our method on all metrics. Furthermore, its qualitative results often suffer from blurry backgrounds, leading to a degradation in overall image quality (Figure~\ref{fig:comparison}).

Furthermore, as depicted in Figure~\ref{fig:multicase}~(a), our method can robustly generate a given instance in diverse backgrounds. Figure~\ref{fig:multicase}~(b) shows our method exhibits strong generalization, generating high-quality images across various poses. Figure~\ref{fig:multicase}~(c) demonstrates the generation diversity of our method.

\begin{figure}[t]
  \centering
   \includegraphics[width=\linewidth]{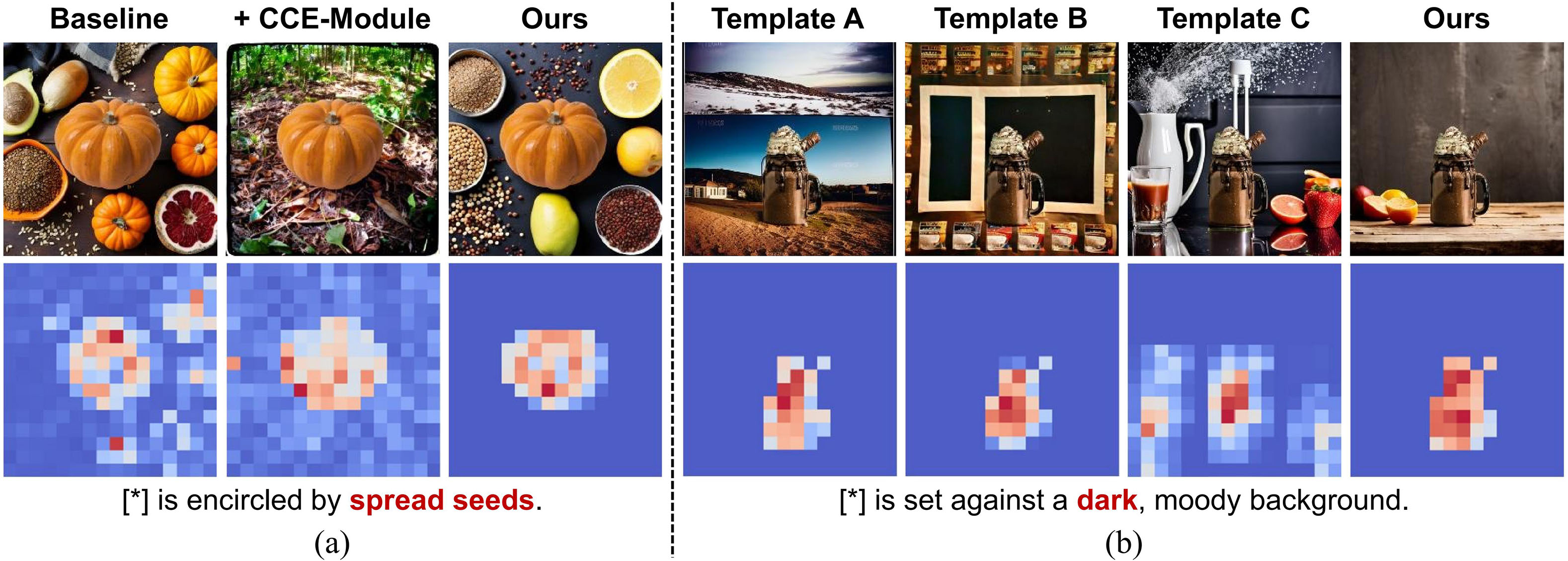}

   \caption{Qualitative comparison for the ablation study on the proposed module and loss (a), and for different prompt templates (b).}
   \label{fig:ablationstudy}
\end{figure}

\begin{figure}[t]
  \centering
   \includegraphics[width=\linewidth]{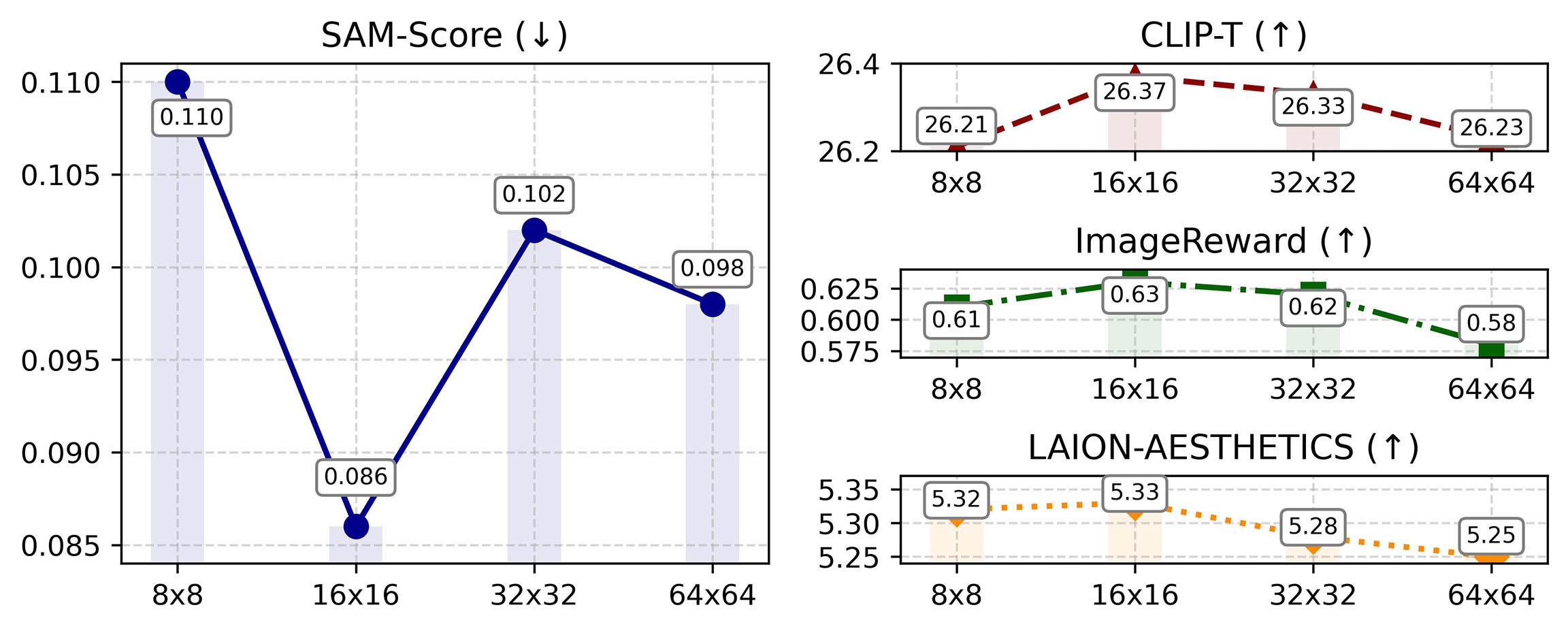}

   \caption{Quantitative analysis across different cross-attention map resolutions. The x-axis indicates the map resolution, and the y-axis represents the corresponding metric score.}
   \label{fig:Attnmaps}
\end{figure}

\subsection{Ablation Study}

\noindent
\textbf{Ablation Study for CCE-Module:} As shown in Table~\ref{tab:ablation_study}, adding the CCE-Module to the Baseline~\cite{module_based_Controlnet} improves the SAM score by 0.109, though at the cost of a decline in other metrics. We attribute this trade-off to potentially inaccurate customized embeddings generated by the CCE-Module, which adversely affect background generation and can thus negatively impact overall image quality.

\noindent
\textbf{Ablation Study for Instance-Aware Loss:} After incorporating this loss, our full model (\textbf{Ours}) demonstrates improvements across all metrics. This indicates that the loss provides a corrective signal to the optimization of CCE-Module, yielding more accurate customized concept embeddings. Consequently, this synergy not only reduces artifacts but also enhances overall image quality, as qualitatively demonstrated in Figure~\ref{fig:ablationstudy}~(a). 

\subsection{Study of Text Templates and Hyperparameters}
\noindent
\textbf{Comparison of Text Templates:} As shown in Table~\ref{tab:template_comparison} and Figure~\ref{fig:ablationstudy}~(b), our proposed template is superior. In contrast to Templates A and B, which achieve low SAM scores at the expense of image quality, and Template C, which suffers from significant artifacts, our template successfully maintains a low artifact level while delivering substantially higher image quality, achieving the optimal trade-off.

\noindent
\textbf{The Selection of Cross-Attention Maps:} As illustrated in Figure~\ref{fig:Attnmaps}, selecting a $16 \times 16$ resolution for the cross-attention maps yields the best overall performance. This represents an optimal trade-off: lower resolutions (e.g., $8 \times 8$) lack sufficient spatial detail for accurate localization, while higher resolutions (e.g., $32 \times 32$ and $64 \times 64$) introduce complex semantic details that can conflict with the loss's objective, making the $16 \times 16$ resolution the optimal choice.

\section{Conclusion}
\label{sec:Conclusion}
In this work, to address the artifact issue in the FCO task, we propose a framework with a CCE-Module and the Instance-Aware Loss for precise instance-related customized concept embeddings, and a Semantic-Preserving Prompt Template to maintain textual semantics. Our method significantly reduces artifacts and improves overall image quality when combined with ControlNet, and can also be integrated with other methods to boost their FCO performance.

\vfill\pagebreak

\noindent\textbf{Acknowledgements.} This work was supported in part by the National Natural Science Foundation of China under Grant No.  62473290, the Shanghai Municipal Science and Technology Commission under Grant No. 25511102800, the National Key R\&D Program of China under Grant No. 2024YFB3311801, and the Shanghai Municipal Commission of Economy and Information Technology under Grant No. 2024-GZL-RGZN-01008.
\bibliographystyle{IEEEbib}
\bibliography{ref_final}

\end{document}